\documentclass{article}

\PassOptionsToPackage{numbers, compress}{natbib}
\usepackage[final]{nips_2016}

\usepackage[utf8]{inputenc} 
\usepackage[T1]{fontenc}    
\usepackage{hyperref}       
\usepackage{url}            
\usepackage{booktabs}       
\usepackage{amsfonts}       
\usepackage{nicefrac}       
\usepackage{microtype}      
\usepackage{graphicx}
\usepackage{amsmath,amsthm,amssymb}
\usepackage{mathrsfs}

\title{Deep FisherNet for Object Classification}

\author{
  Peng Tang$^{\dag}$, Xinggang Wang$^{\dag}$, Baoguang Shi$^{\dag}$, Xiang Bai$^{\dag}$, Wenyu Liu$^{\dag}$, Zhuowen Tu$^{\ddag}$ \\
  $^{\dag}$School of EIC, Huazhong University of Science and Technology\\
  $^{\ddag}$Dept. of CogSci and Dept. of CSE, University of California, San Diego\\
  \texttt{$\{$pengtang,xgwang,shibaoguang,xbai,liuwy$\}$@hust.edu.cn, ztu@ucsd.edu} \\
}

\begin{document}

\maketitle

\begin{abstract}
Despite the great success of convolutional neural networks (CNN) for the image classification task on datasets like Cifar and ImageNet, CNN's representation power is still somewhat limited in dealing with object images that have large variation in size and clutter,
where Fisher Vector (FV) has shown to be an effective encoding strategy.
FV encodes an image by aggregating local descriptors with a universal generative Gaussian Mixture Model (GMM). 
FV however has limited learning capability and its parameters are mostly fixed after constructing the codebook.
To combine together the best of the two worlds, we propose in this paper a neural network structure with FV layer being part of an end-to-end trainable system that is differentiable; we name our network \emph{FisherNet} that is learnable using back-propagation.
Our proposed FisherNet combines convolutional neural network training and Fisher Vector encoding in a single end-to-end structure.
We observe a clear advantage of FisherNet over plain CNN and standard FV in terms of both classification accuracy and computational efficiency on the challenging PASCAL VOC object classification task.
 
\end{abstract}

\vspace{-0.37cm}
\section{Introduction}
\label{sec:intro}
\vspace{-0.37cm}


Convolutional Neural Networks (CNNs)~\cite{Ref:LeCun1989,Ref:Krizhevsky2012} have led to leap-forward in a large number of computer vision applications. On the task of large scale image classification, particularly ImageNet~\cite{Ref:Deng2009}, CNNs-family models have been dominating. CNNs are able to automatically learn rich hierachical features from input images.

\begin{figure}[t]
  \centerline{
    \includegraphics[width=0.8\textwidth]{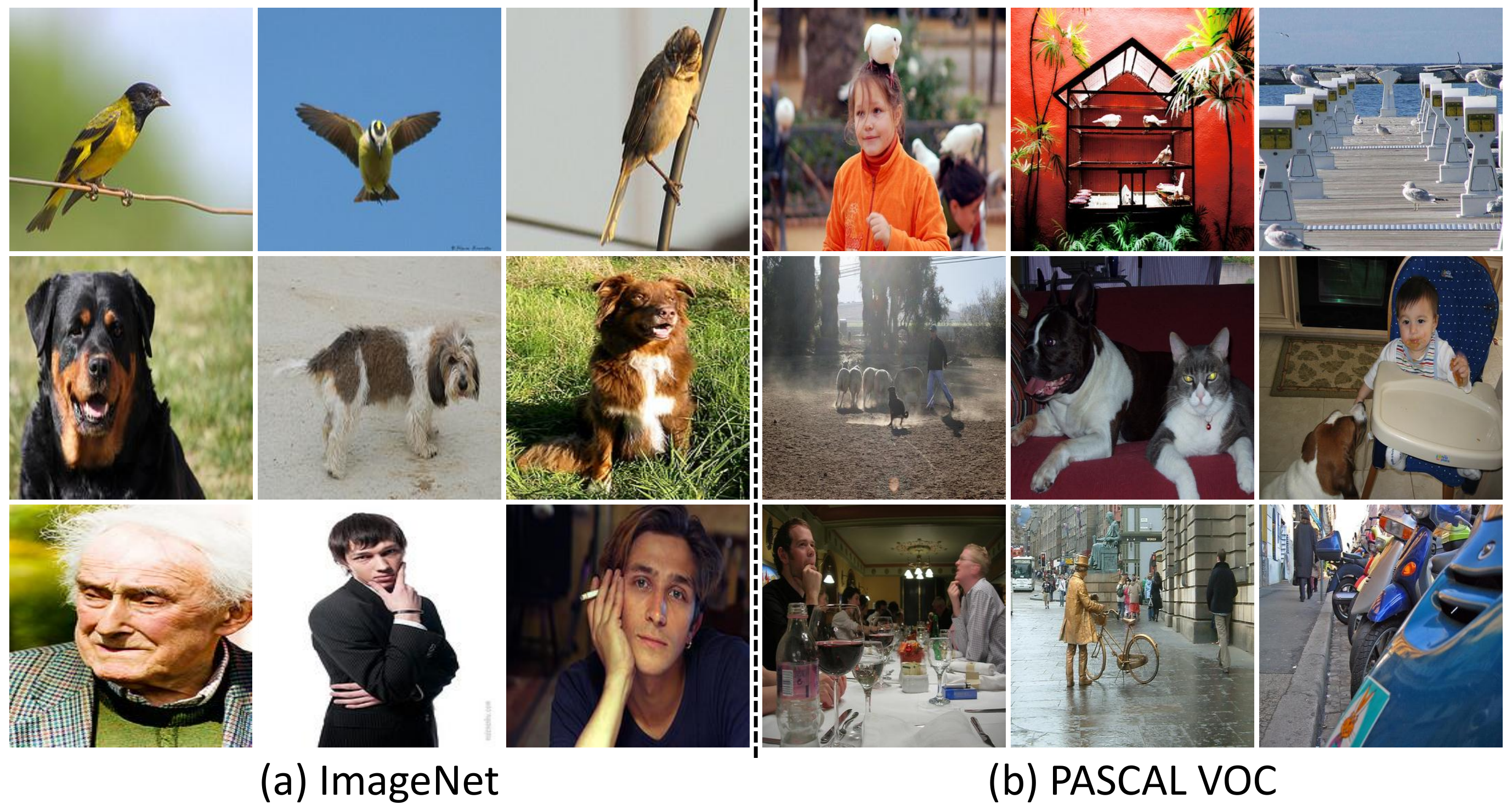}
  }
  \caption{Randomly selected images from (a) ImageNet and (b) PASCAL VOC. 
  From top row to down, images are from bird, dog, and person. 
  Images from PASCAL VOC are more complex.}
  \label{fig:datasets}
\end{figure}

However, for images dataset like PASCAL VOC~\cite{Ref:Everingham2010} where objects have a large variation in shape, size, and clutter, directly adopting CNN does not produce satisfactory results: state-of-the-art results for PASCAL VOC object classification are obtained  with 
Bag of Visual Words (BoVW) on top of the CNN features that are learned separately.
As shown in Fig.~\ref{fig:datasets} (a), ImageNet mainly consists of \emph{iconic-object images}, i.e. single large objects in the canonical perspective are located in the center of these images. Compared with ImageNet, structures of PASCAL VOC images tend to be much more complex. Objects have large variations in location, scale, layout, and appearance; The backgrounds are cluttered; There tends to be multiple objects in an image. A standard pipeline includes (1) local feature extraction using off-the-shelf CNNs that are pretrained on ImageNet; (2) sparse coding~\cite{Ref:Yang2009} or Fisher Vector~\cite{Ref:Sanchez2013} adopted to aggregate local features into a global, fixed-dimensional image representation; (3)  classification on the encoded feature space. 
These specific approaches often produce results that are much better than those by plain CNN~\cite{Ref:Liu2014,Ref:Cimpoi2015}.



Due to the complexity of BoVW based methods, most previous works extract representations with a standalone module which cannot be trained together with other modules. 
Consequently, former modules in their algorithm pipelines such as the CNN feature extractor does not receive error differentials from latter ones, and thus cannot be finetuned. 
This has negative impacts on the overall performance. 
In one particular aspect, CNN features are learned from ImageNet, whose object distribution is quite different from that in PASCAL VOC. 
Without finetuning, the features are likely to be less effective. 


In this paper, we propose \emph{FisherNet}, an end-to-end trainable neural network which takes advantage of both CNN features and the powerful Fisher Vector (FV)~\cite{Ref:Sanchez2013} encoding method. 
FisherNet densely extracts local features at multiple scales, and aggregates them with FV.
FV encodes and aggregates local descriptors with a universal generative Gaussian Mixture Model (GMM). We model this process into a learnable module, which we call \emph{Fisher Layer}. The Fisher Layer allows back-propagating error differentials as well as optimizing the FV codebook, eventually making the whole network trainable end-to-end.

Moreover, FisherNet learns and extracts local patch features with great efficiency. 
Inspired by the recent success of fast object detection algorithms such as SPPnet~\cite{Ref:He2015} and Fast R-CNN~\cite{Ref:Girshick2015}, we share the computation of feature extraction among patches.

Experiments show that our FisherNet significantly boosts the performance of an untrained FV, and achieves highly competitive performance on the PASCAL VOC object classification task.
In testing, a FisherNet with the AlexNet~\cite{Ref:Krizhevsky2012} takes only $0.3$s to classify one image, and $0.8$s with the VGG16~\cite{Ref:Simonyan2015}, both over $10\times$ faster than the previous state-of-the-art method HCP~\cite{Ref:Wei2015}.


\vspace{-0.37cm}
\section{Related Work}
\label{sec:related_work}
\vspace{-0.37cm}

Bag of Visual Words (BoVW) based image representation is one of the most popular methods in computer vision community, especially for image classification~\cite{Ref:Yang2009,Ref:Wang2013,Ref:Doersch2013,Ref:Sanchez2013}. 
BoVW has been widely applied for their robustness, especially to object deformation, translation, and occlusion. 
Fisher Vector (FV)~\cite{Ref:Sanchez2013} is one of the most powerful BoVW based representation methods. It has achieved many state-of-the-art performance on image classification. 
The traditional FV uses hand-crafted descriptors like SIFT~\cite{Ref:Lowe2004} as patch features, and learns FV parameters by Gaussian Mixture Models (GMM), which is not trainable for both patch features and FV parameters.

Recently, inspired by the great success of deep CNN on image classification~\cite{Ref:Krizhevsky2012,Ref:Simonyan2015},
many attempts have been made to combine CNN and FV.
Liu et al.~\cite{Ref:Liu2014} extract activations from the first fully connected layer for patch features, and use Sparse Coding based FV for object, scene, and fine-grained recognition.
Cimpoi et al.~\cite{Ref:Cimpoi2015} extract outputs of the last convolutional layer of a CNN as input descriptors, and use FV to represent images for texture recognition, object and scene classification.
Dixit et al.~\cite{Ref:Dixit2015} combine activations from the last fully connected layer and FV for scene classification. 
They all show better results than plain CNN.
But they simply use CNN to extract patch features, and FV parameters are also not trainable, both of which may limit the performance of their methods.

There are many researchers also trying to improve the FV~\cite{Ref:Sydorov2014,Ref:Simonyan2013}. 
Sydorov et al.~\cite{Ref:Sydorov2014} extract patch features by hand-crafted SIFT, and learn parameters of FV and SVM jointly for object classification.
They choose an alternative strategy to optimize FV and SVM parameters, that is, fixing the FV parameters and train the SVM, fixing the SVM parameters and train the FV.
But using SIFT makes it impossible to learn patch features end-to-end.
Meanwhile, the alternative optimization is incompatible with the gradient descend optimization adopted by CNN.
Different from their method, we decompose FV into a series of network layers and insert them to a CNN, and learn both patch features and FV parameters in an end-to-end manner, by standard back-propagation.
As we will show in Section~\ref{sec:discu}, learning parameters of patch features and FV end-to-end outperforms only learning FV parameters by a large margin.
Simonyan et al.~\cite{Ref:Simonyan2013} also propose a ``Fisher Network'' by stacking FVs on the top of each other. However, the network depends on hand-crafted descriptors, and parameters of FV are also fixed upon constructing the codebook.

Recently, a ``NetVLAD'' framework presented by Arandjelovi{\'c} et al.~\cite{Ref:Arandjelovic2016} develop a VLAD layer for deep networks.
They choose outputs from the last convolutional layer as input descriptors, followed by a VLAD layer, which also learns all parameters of patch features and VLAD end-to-end.
But notice that VLAD is just a simplified version of FV~\cite{Ref:Jegou2012,Ref:Sanchez2013}.
It is more difficult to embed FV into CNN frameworks.
Meanwhile, VLAD and NetVLAD are only able to capture first-order statistics of data, while FV and FisherNet capture both first- and second-order statics.
So in many applications especially for image classification, FV is more suitable\cite{Ref:Cimpoi2015,Ref:Dixit2015}.
Moreover, as receptive field sizes of convolutional layers are fixed, the patches from the last convolutional layer are only with one scale.
We share the computation of convolutional layer for different patches, and use Spatial Pyramid Pooling (SPP) layer~\cite{Ref:He2015} to generate patch features, making it possible to extract features from patches at multiple scales.

\vspace{-0.37cm}
\section{The Architecture of Deep FisherNet}
\label{sec:deep_fishernet}
\vspace{-0.37cm}

\begin{figure}[t]
  \centerline{
    \includegraphics[width=13.7cm]{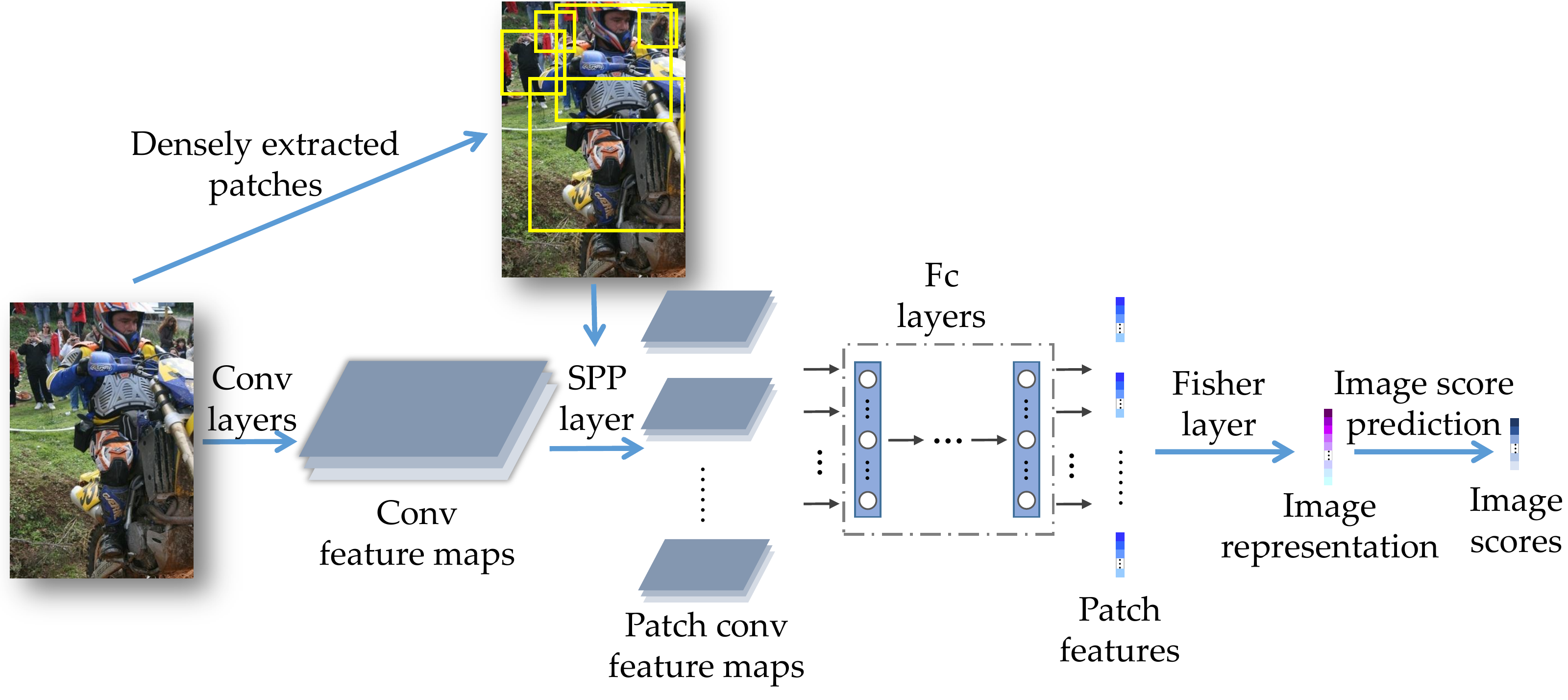}
  }
  \caption{The architecture of deep FisherNet. 
  Some patches are first densely extracted from an input image, meanwhile, the image is passed through several conv layers to produce a conv feature map. 
  Then a SPP layer is employed to project each patch to a fixed-size feature map, and be fed into some fc layers to output some patch features. 
  The Fisher Layer is implemented to aggregate these features into a representation, as shown in Fig.~\ref{fig:fisher_layer}. 
  At last, this representation is used for classification.}
  \label{fig:fishernet_architecture}
\end{figure}

The architecture of our FisherNet is shown in Fig.~\ref{fig:fishernet_architecture}. 
Given an input image, we first densely extract image patches at multiple scales.
These patches cover objects with different locations and scales.
The image is passed through several convolutional (conv) layers, finally resulting in a conv feature map, and the size of this map varies as the size of input image changes.
Then, a SPP layer~\cite{Ref:He2015} is employed to generate a fixed-size conv feature map for every patch, since the following fully connected (fc) layers require fixed-length input.
The fc layers accept each of the feature maps and output a patch feature vector correspondingly. It eventually outputs a collection of patch features from the set of feature maps.
Following that, the collection is fed into our proposed Fisher Layer, which aggregates patch features into an orderless fixed-length image representation.
At last, this representation is used for classification. 
In this section, we will describe our whole FisherNet architecture for object classification.

\vspace{-0.25cm}
\subsection{Pre-trained CNN Models}
\label{sec:pretrained_cnn}
\vspace{-0.25cm}

As the number of training images is limited, it seems unpractical to train a randomly initialized CNN model on target dataset directly. 
It is more advisable to finetune a CNN model trained on large datasets like ImageNet~\cite{Ref:Deng2009}. 
Here we choose two CNN models AlexNet~\cite{Ref:Krizhevsky2012} and VGG16~\cite{Ref:Simonyan2015}, both have several conv layers with some max-pooling layers and three fc layers.

\vspace{-0.25cm}
\subsection{SPP Layer}
\vspace{-0.25cm}

One way to generate patch features is feeding each patch into CNN models separately. 
But this tends to be time-consuming since it does not share the computation of overlapping patches.
Here we adopt the SPP layer method in SPPnet~\cite{Ref:He2015} and Fast RCNN~\cite{Ref:Girshick2015}. 
After obtaining the feature map of input image, the SPP layer is employed to project each patch to a fixed-size feature map. 
Specifically, the last max-pooling layer of the original CNN is replaced by the SPP layer. 
For patch $R_{i}$, its output feature map $y_{i}$ can be acquired by Eq.~(\ref{equ:SPP}), 
where $x_{k}$ is the $k$-th activation into the SPP layer, $R_{ij}$ is the $j$-th sub-window of $R_{i}$, and $y_{ij}$ is the output from $R_{ij}$. 
The number of subwindows depends on the original CNN model (e.g., $6\times6$ for AlexNet~\cite{Ref:Krizhevsky2012} and $7\times7$ for VGG16~\cite{Ref:Simonyan2015}). 
After the SPP layer, each patch feature map is passed to following fc layer to produce patch features.

\begin{equation}
    \label{equ:SPP}
    y_{ij} = \mathop{\max} \limits_{k \in R_{ij}} x_{k}.
\end{equation}

\vspace{-0.25cm}
\subsection{Fisher Layer}
\label{sec:fisher_layer}
\vspace{-0.25cm}

After generating patch features, we aggregate them into an image representation. 
This is implemented by our Fisher Layer.  
In this subsection, we will first review the FV~\cite{Ref:Sanchez2013} for image classification briefly, then we present the proposed Fisher Layer of FisherNet.

\vspace{-0.25cm}
\subsubsection{Fisher Vector}
\vspace{-0.25cm}

Let ${\cal X}=\{X_{i}\}_{i=1}^{N}$ be a set of images. Their patch features are $X_{i} = \{\mathbf{x}_{ij}\}_{j=1}^{m_{i}}, \mathbf{x}_{ij} \in \mathbb{R}^{D \times 1}$, where $N$ and $m_{i}$ are the number of images and the number of patches for image $X_{i}$ respectively. 
A $K$-component Gaussian Mixture Model (GMM) $u_{\lambda}(\mathbf{x}) = \mathop \sum \limits_{k=1}^{K} \omega_{k} u_{k}(\mathbf{x})$ is used as probability density function of $\mathbf{x}_{ij}$. 
Let $\lambda = \{\omega_{k}, \mathbf{\mu}_{k}, \mathbf{\Sigma}_{k}, k = 1, 2, ..., K\}, \omega_{k} \in \mathbb{R}, \mathbf{\mu}_{k} \in \mathbb{R}^{D \times 1}, \mathbf{\Sigma}_{k} \in \mathbb{R}^{D \times D}$ be parameters of GMM, 
where $\omega_{k}, \mathbf{\mu}_{k}, \mathbf{\Sigma}_{k}$ are weight, mean vector, covariance matrix of $k$-th GMM component, respectively. 
Then $k$-th Gaussian distribution $u_{k}(\mathbf{x})$ can be written as 

\begin{equation}
    \label{equ:gaussian}
    u_{k}(\mathbf{x}) = \frac{1}{(2 \pi)^{D/2} \vert \mathbf{\Sigma}_{k} \vert^{1/2}} \exp \{ -\frac{1}{2} (\mathbf{x} - \mathbf{\mu}_{k})^{T} \mathbf{\Sigma}_{k}^{-1} (\mathbf{x} - \mathbf{\mu}_{k}) \}.
\end{equation}

Notice that covariance matrix $\mathbf{\Sigma}_{k}$ is restricted to be a diagonal matrix in FV~\cite{Ref:Sanchez2013}, 
i.e., $\mathbf{\Sigma}_{k} = diag(\mathbf{\sigma}_{k}^{2}), \mathbf{\sigma}_{k} \in \mathbb{R}^{D \times 1}$. 
For any patch feature $\mathbf{x}_{ij}$, we define a vector $\mathbf{\varphi}(\mathbf{x}_{ij}) = [{\mathscr{G}_{\mathbf{\mu}_{1}}^{\mathbf{x}_{ij}}}^{T}, ..., {\mathscr{G}_{\mathbf{\mu}_{K}}^{\mathbf{x}_{ij}}}^{T}, {\mathscr{G}_{\mathbf{\sigma}_{1}}^{\mathbf{x}_{ij}}}^{T}, ..., {\mathscr{G}_{\mathbf{\sigma}_{K}}^{\mathbf{x}_{ij}}}^{T}]^{T} \in \mathbb{R}^{2KD \times 1}$, 
where its subvector $\mathscr{G}_{\mathbf{\mu}_{k}}^{\mathbf{x}_{ij}}$ and $\mathscr{G}_{\mathbf{\sigma}_{k}}^{\mathbf{x}_{ij}}$ are as follows

\begin{equation}
    \label{equ:fv_mu}
    \mathscr{G}_{\mathbf{\mu}_{k}}^{\mathbf{x}_{ij}} = \frac{1}{\sqrt{\omega_{k}}} 
    \gamma_{j}(k) (\frac{\mathbf{x}_{ij} - \mathbf{\mu}_{k}}{\mathbf{\sigma}_{k}}),
\end{equation}

\begin{equation}
    \label{equ:fv_sigma}
    \mathscr{G}_{\mathbf{\sigma}_{k}}^{\mathbf{x}_{ij}} = \frac{1}{\sqrt{\omega_{k}}} 
    \gamma_{j}(k) \frac{1}{\sqrt{2}} [\frac{(\mathbf{x}_{ij} - \mathbf{\mu}_{k})^{2}}{\mathbf{\sigma}_{k}^{2}} - 1].
\end{equation}
The $\gamma_{j}(k)$ in Eq.~(\ref{equ:fv_mu}) and (\ref{equ:fv_sigma}) is posterior probability as in Eq.~(\ref{equ:gamma}). 
Then the FV $\mathbf{\phi}(X_{i})$ of image $X_{i}$ is the mean-pooling of all patch representations in $X_{i}$, 
i.e., $\mathbf{\phi}(X_{i}) = \frac{1}{m_{i}} \mathop \sum \limits_{j=1}^{m_{i}} \mathbf{\varphi}(\mathbf{x}_{ij})$.

\begin{equation}
    \label{equ:gamma}
    \gamma_{j}(k) = \frac{\omega_{k} u_{k}(\mathbf{x}_{ij})}{\mathop \sum \limits_{n=1}^{K} \omega_{n} u_{n}(\mathbf{x}_{ij})}.
\end{equation}

\vspace{-0.25cm}
\subsubsection{The Architecture of Fisher Layer}
\label{sec:archi_fisher_layer}
\vspace{-0.25cm}

The parameters of traditional FV are fixed once codebook is constructed.
Fisher Layer, however, has learnable parameters after the codebook construction, thus can be trained jointly with CNN.
To achieve this, we first make two simplification to the original FV:
1) We drop the weight $\omega_{k}$, which assumes all GMM components have equal weights;
2) We simplify $u_{k}(\mathbf{x})$ to be the form in Eq.~(\ref{equ:gaussian_simple}), which is similar to covariance matrices share the same determinants.
Despite the small differences from the original FV, our simplified FV still inherits the superiority of capturing first- and second-order statistics.
We will also show in Section~\ref{sec:discu} that even with these simplifications, our FisherNet still achieves better performance than the traditional FV.

\begin{equation}
    \label{equ:gaussian_simple}
    u_{k}(\mathbf{x}) = \frac{1}{(2 \pi)^{D/2}} \exp \{ -\frac{1}{2} (\mathbf{x} - \mathbf{\mu}_{k})^{T} \mathbf{\Sigma}_{k}^{-1} (\mathbf{x} - \mathbf{\mu}_{k}) \}.
\end{equation}

After these simplifications, the $\gamma_{j}(k)$ can be rewritten as Eq.~(\ref{equ:gamma_simple_final}), with $\mathbf{\Sigma}_{k} = diag(\mathbf{\sigma}_{k}^{2})$ been submitted into Eq.~(\ref{equ:gamma}). 




\begin{equation}
    \label{equ:gamma_simple_final}
    \gamma_{j}(k) = \frac{u_{k}(\mathbf{x}_{ij})}{\mathop \sum \limits_{n=1}^{K}u_{n}(\mathbf{x}_{ij})} 
    = \frac{\exp \{ -\frac{1}{2} (\frac{\mathbf{x}_{ij} - \mathbf{\mu}_{k}}{\mathbf{\sigma}_{k}})^{T} (\frac{\mathbf{x}_{ij} - \mathbf{\mu}_{k}}{\mathbf{\sigma}_{k}}) \}}
    {\mathop \sum \limits_{n=1}^{K} \exp \{ -\frac{1}{2} (\frac{\mathbf{x}_{ij} - \mathbf{\mu}_{n}}{\mathbf{\sigma}_{n}})^{T} (\frac{\mathbf{x}_{ij} - \mathbf{\mu}_{n}}{\mathbf{\sigma}_{n}}) \}}.
\end{equation}

\begin{figure}[t]
  \centerline{
    \includegraphics[width=13.7cm]{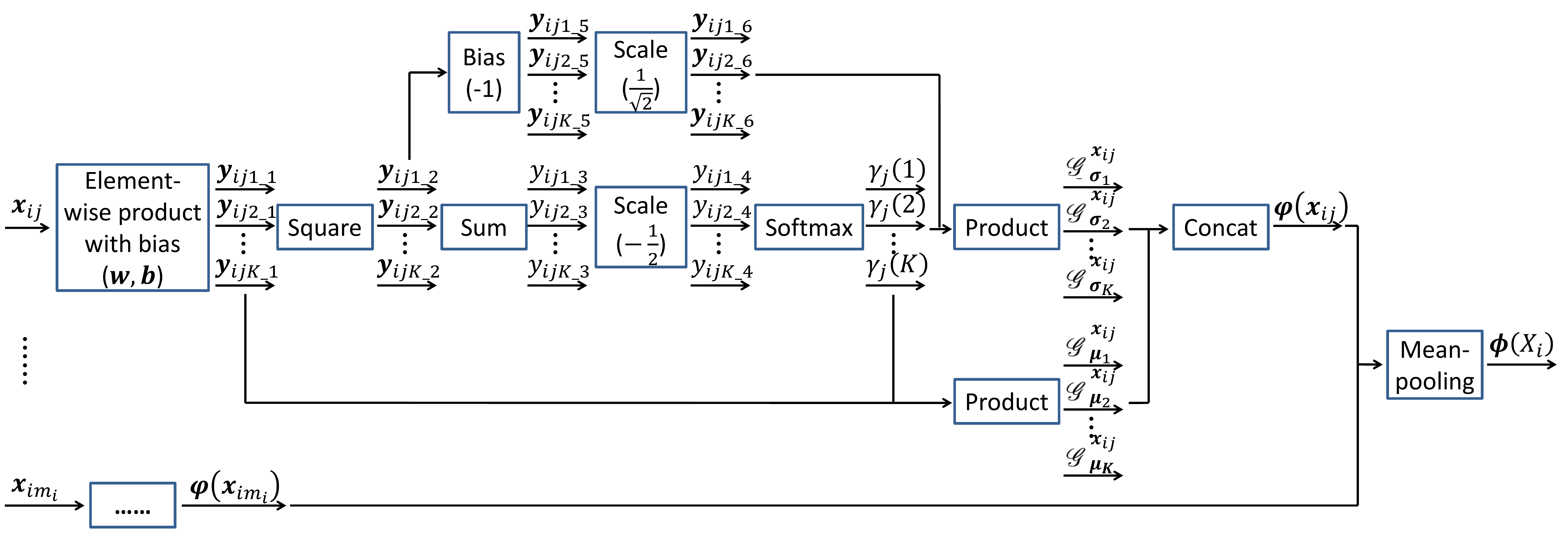}
  }
  \caption{The architecture of Fisher Layer, where $\mathbf{y}_{ijk\_1} = \mathbf{w}_{k} \odot (\mathbf{x}_{ij} + \mathbf{b}_{k})$, $\mathbf{y}_{ijk\_2} = (\mathbf{y}_{ijk\_1}) ^ {2}$, $y_{ijk\_3} = \sum_{d=1}^{D} y_{ijk\_2d} = \mathbf{y}_{ijk\_1}^{T} \mathbf{y}_{ijk\_1}$, ``Scale'' and ``Bias'' are some linear transformation, i.e., $y_{ijk\_4} = -\frac{1}{2} y_{ijk\_3}$ and $\mathbf{y}_{ijk\_5} = \mathbf{y}_{ijk\_2} - 1$.}
  \label{fig:fisher_layer}
\end{figure}

Suppose $\mathbf{w}_{k} = 1 / \mathbf{\sigma}_{k}$ and $\mathbf{b}_{k} = -\mathbf{\mu}_{k}$, the final form of Fisher Layer can be obtained as follows

\begin{equation}
    \label{equ:fv_mu_final}
    \mathscr{G}_{\mathbf{\mu}_{k}}^{\mathbf{x}_{ij}} = \gamma_{j}(k) [\mathbf{w}_{k} \odot (\mathbf{x}_{ij} + \mathbf{b}_{k})],
\end{equation}

\begin{equation}
    \label{equ:fv_sigma_final}
    \mathscr{G}_{\mathbf{\sigma}_{k}}^{\mathbf{x}_{ij}} = \gamma_{j}(k) \frac{1}{\sqrt{2}} [(\mathbf{w}_{k} \odot (\mathbf{x}_{ij} + \mathbf{b}_{k}))^{2} - 1],
\end{equation}

\begin{equation}
    \label{equ:gamma_final}
    \gamma_{j}(k) = \frac{\exp \{- \frac{1}{2} (\mathbf{w}_{k} \odot (\mathbf{x}_{ij} + \mathbf{b}_{k}))^{T} (\mathbf{w}_{k} \odot (\mathbf{x}_{ij} + \mathbf{b}_{k}))\}}
    {\mathop \sum \limits_{n=1}^{K} \exp \{- \frac{1}{2} (\mathbf{w}_{n} \odot (\mathbf{x}_{ij} + \mathbf{b}_{n}))^{T} (\mathbf{w}_{n} \odot (\mathbf{x}_{ij} + \mathbf{b}_{n}))\}},
\end{equation}
where $\odot$ is an element-wise product operation. 
Notice that Eq.~(\ref{equ:gamma_final}) is just a softmax function, $\mathbf{w}_{k}$ and $\mathbf{b}_{k}$ are sets of learnable parameters for each GMM component $k$. 
We can observe that Eq.~(\ref{equ:fv_mu_final}), (\ref{equ:fv_sigma_final}), and (\ref{equ:gamma_final}) share the same computation part $\mathbf{w}_{k} \odot (\mathbf{x}_{ij} + \mathbf{b}_{k})$, which is obviously differentiable. 
Meanwhile, others are just some linear or square operations, so we can derive all parameters via back-propagation. 
To make the Fisher Layer more clear, we also show the architecture in Fig.~\ref{fig:fisher_layer}.

\vspace{-0.25cm}
\subsection{Loss Function}
\vspace{-0.25cm}

In above subsections, we describe some important parts of our FisherNet. 
In this subsection, we will introduce our loss. 
Since we are focusing on object classification, which may have multiple different objects in the same image, the popular softmax loss is not suitable. 
Here we choose the muti-class sigmoid cross entropy loss. 
Specifically, suppose the predicted score vector of image $X_{i}$ is $\mathbf{s}_{i} = [s_{i1}, ..., s_{iC}]^{T} \in \mathbb{R}^{C \times 1}$; label vector is $\mathbf{y}_{i} = [y_{i1}, ..., y_{iC}]^{T} \in \mathbb{R}^{C \times 1}$, where $C$ is the number of classes, $y_{ic} = 0$ if $X_{i}$ has object $c$ or $y_{ic} = 0$ otherwise. 
Then the loss function will be Eq.~(\ref{equ:loss}), where $\sigma (x)$ is the sigmoid function, and it can be described as $\sigma (x) = 1 / (1 + \exp(-x))$. 
Actually, the loss can be changed for different tasks. 

\begin{equation}
\label{equ:loss}
    Loss(\mathbf{s}_{i}, \mathbf{y}_{i}) = -\mathop{\sum} \limits_{c=1}^{C} \{y_{ic}\log\, \sigma (s_{ic}) + (1 - y_{ic})\log\,(1 - \sigma (s_{ic}))\}.
\end{equation}

\begin{table}[t]
  \caption{Object classification results (AP in $\%$) for different methods on PASCAL VOC 2007 test set.}
  \label{table:pascal_2007}
  \centering
  \begin{tabular}{p{3.2cm} | p{0.52cm} p{0.52cm} p{0.52cm} p{0.52cm} p{0.52cm} p{0.52cm} p{0.52cm} p{0.52cm} p{0.52cm} p{0.52cm} | p{0.52cm}}
    \toprule
    method & aero & bike & bird & boat & bottle & bus & car & cat & chair & cow &\\
    \midrule
    SCFVC~\cite{Ref:Liu2014} & $89.5$ & $84.1$ & $83.7$ & $83.7$ & $43.9$ & $76.7$ & $87.8$ & $82.5$ & $60.6$ & $69.6$ &\\
    Cimpoi et al.~\cite{Ref:Cimpoi2015} & - & - & - & - & - & - & - & - & - & - &\\
    VGG16-19-SVM~\cite{Ref:Simonyan2015} & - & - & - & - & - & - & - & - & - & - &\\
    HCP-VGG16~\cite{Ref:Wei2015} & $\mathbf{98.6}$ & $\mathbf{97.1}$ & $\mathbf{98.0}$ & $95.6$ & $75.3$ & $\mathbf{94.7}$ & $95.8$ & $\mathbf{97.3}$ & $73.1$ & $\mathbf{90.2}$ &\\
    \midrule
    FisherNet-AlexNet & $94.6$ & $90.4$ & $90.3$ & $89.1$ & $54.3$ & $82.1$ & $93.0$ & $89.8$ & $66.6$ & $78.2$ &\\
    FisherNet-VGG16 & $98.2$ & $96.7$ & $97.5$ & $\mathbf{95.9}$ & $\mathbf{78.0}$ & $92.8$ & $\mathbf{96.4}$ & $96.9$ & $\mathbf{75.7}$ & $88.9$ &\\
    \bottomrule
  \end{tabular}
  \begin{tabular}{p{3.2cm} | p{0.52cm} p{0.52cm} p{0.52cm} p{0.52cm} p{0.52cm} p{0.52cm} p{0.52cm} p{0.52cm} p{0.52cm} p{0.52cm} | p{0.52cm}}
    \toprule
    method & table & dog & horse & mbike & persn & plant & sheep & sofa & train & tv & mAP\\
    \midrule
    SCFVC~\cite{Ref:Liu2014} & $72.0$ & $77.1$ & $88.7$ & $82.1$ & $94.4$ & $56.8$ & $71.4$ & $67.7$ & $90.9$ & $75.0$ & $76.9$\\
    Cimpoi et al.~\cite{Ref:Cimpoi2015} & - & - & - & - & - & - & - & - & - & - & $88.6$\\
    VGG16-19-SVM~\cite{Ref:Simonyan2015} & - & - & - & - & - & - & - & - & - & - & $89.7$\\
    HCP-VGG16~\cite{Ref:Wei2015} & $80.0$ & $\mathbf{97.3}$ & $96.1$ & $\mathbf{94.9}$ & $96.3$ & $78.3$ & $\mathbf{94.7}$ & $76.2$ & $97.9$ & $\mathbf{91.5}$ & $90.9$\\
    \midrule
    FisherNet-AlexNet & $78.1$ & $88.0$ & $92.2$ & $88.3$ & $96.9$ & $67.2$ & $81.7$ & $74.1$ & $94.7$ & $78.0$ & $83.4$ \\
    FisherNet-VGG16 & $\mathbf{87.4}$ & $96.8$ & $\mathbf{97.0}$ & $94.5$ & $\mathbf{99.0}$ & $\mathbf{80.5}$ & $92.7$ & $\mathbf{81.0}$ & $\mathbf{98.1}$ & $90.6$ & $\mathbf{91.7}$\\
    \bottomrule
  \end{tabular}
\end{table}

\begin{table}[t]
  \caption{Object classification results (AP in $\%$) for different methods on PASCAL VOC 2012 test set.}
  \label{table:pascal_2012}
  \centering
  \begin{tabular}{p{3.2cm} | p{0.52cm} p{0.52cm} p{0.52cm} p{0.52cm} p{0.52cm} p{0.52cm} p{0.52cm} p{0.52cm} p{0.52cm} p{0.52cm} | p{0.52cm}}
    \toprule
    method & aero & bike & bird & boat & bottle & bus & car & cat & chair & cow &\\
    \midrule
    Oquab et al.~\cite{Ref:Oquab2015} & $96.7$ & $88.8$ & $92.0$ & $87.4$ & $64.7$ & $91.1$ & $87.4$ & $94.4$ & $74.9$ & $89.2$ &\\
    VGG16-19-SVM~\cite{Ref:Simonyan2015} & $99.1$ & $89.1$ & $96.0$ & $94.1$ & $74.1$ & $92.2$ & $85.3$ & $97.9$ & $79.9$ & $92.0$ &\\
    HCP-VGG16~\cite{Ref:Wei2015} & $99.1$ & $\mathbf{92.8}$ & $\mathbf{97.4}$ & $94.4$ & $79.9$ & $\mathbf{93.6}$ & $89.8$ & $98.2$ & $78.2$ & $\mathbf{94.9}$ &\\
    \midrule
    FisherNet-AlexNet & $96.6$ & $83.3$ & $90.4$ & $87.8$ & $59.4$ & $88.3$ & $83.4$ & $93.6$ & $72.4$ & $76.9$ &\\
    FisherNet-VGG16 & $\mathbf{99.2}$ & $92.5$ & $96.8$ & $\mathbf{94.4}$ & $\mathbf{81.0}$ & $93.2$ & $\mathbf{92.3}$ & $\mathbf{98.2}$ & $\mathbf{82.9}$ & $94.3$ &\\
    \bottomrule
  \end{tabular}
  \begin{tabular}{p{3.2cm} | p{0.52cm} p{0.52cm} p{0.52cm} p{0.52cm} p{0.52cm} p{0.52cm} p{0.52cm} p{0.52cm} p{0.52cm} p{0.52cm} | p{0.52cm}}
    \toprule
    method & table & dog & horse & mbike & persn & plant & sheep & sofa & train & tv & mAP\\
    \midrule
    Oquab et al.~\cite{Ref:Oquab2015} & $76.3$ & $93.7$ & $95.2$ & $91.1$ & $97.6$ & $66.2$ & $91.2$ & $70.0$ & $94.5$ & $83.7$ & $86.3$\\
    VGG16-19-SVM~\cite{Ref:Simonyan2015} & $\mathbf{83.7}$ & $97.5$ & $96.5$ & $94.7$ & $97.1$ & $63.7$ & $93.6$ & $75.2$ & $97.4$ & $87.8$ & $89.3$\\
    HCP-VGG16~\cite{Ref:Wei2015} & $79.8$ & $\mathbf{97.8}$ & $97.0$ & $93.8$ & $96.4$ & $\mathbf{74.3}$ & $94.7$ & $71.9$ & $96.7$ & $88.6$ & $90.5$\\
    \midrule
    FisherNet-AlexNet & $75.3$ & $91.2$ & $89.5$ & $89.1$ & $96.8$ & $61.1$ & $83.3$ & $66.7$ & $93.1$ & $80.9$ & $83.0$\\
    FisherNet-VGG16 & $82.2$ & $97.4$ & $\mathbf{97.3}$ & $\mathbf{95.9}$ & $\mathbf{98.7}$ & $72.9$ & $\mathbf{95.1}$ & $\mathbf{77.7}$ & $\mathbf{97.5}$ & $\mathbf{90.8}$ & $\mathbf{91.5}$\\
    \bottomrule
  \end{tabular}
\end{table}


\begin{table}[t]
  \caption{The improvement of our FisherNet on PASCAL VOC 2007 and 2012 using AlexNet (mAP in $\%$). 
  The details of CNN-finetune, CNN-FV, CNN-FL, and FisherNet are described in Section~\ref{sec:discu}.}
  \label{table:comparsion}
  \centering
  \begin{tabular}{c | c c c c}
    \toprule
    dataset & CNN-finetune & CNN-FV & CNN-FL & FisherNet\\
    \midrule
    PASCAL VOC 2007 & $76.9$ & $80.5$ & $81.8$ & $\mathbf{83.4}$\\
    \midrule
    PASCAL VOC 2012 & $76.3$ & $80.1$ & $81.3$ & $\mathbf{83.0}$\\
    \bottomrule
  \end{tabular}
\end{table}

\begin{table}[t]
  \caption{The comparison of our FisherNet and traiditional FV on PASCAL VOC 2007 using AlexNet (AP in $\%$). 
  Our learning strategy can boost the performance in all cases.}
  \label{table:improvement}
  \centering
  \begin{tabular}{p{2cm} | p{0.52cm} p{0.52cm} p{0.52cm} p{0.52cm} p{0.52cm} p{0.52cm} p{0.52cm} p{0.52cm} p{0.52cm} p{0.52cm} | p{0.52cm}}
    \toprule
    & aero & bike & bird & boat & bottle & bus & car & cat & chair & cow &\\
    \midrule
    CNN-FV & $91.6$ & $86.5$ & $86.9$ & $85.4$ & $50.0$ & $80.2$ & $89.7$ & $86.5$ & $61.3$ & $76.6$\\
    FisherNet & $\mathbf{94.6}$ & $\mathbf{90.4}$ & $\mathbf{90.3}$ & $\mathbf{89.1}$ & $\mathbf{54.3}$ & $\mathbf{82.1}$ & $\mathbf{93.0}$ & $\mathbf{89.8}$ & $\mathbf{66.6}$ & $\mathbf{78.2}$ &\\
    \midrule
    Improvement & $+3.0$ & $+3.9$ & $+3.4$ & $+3.7$ & $+4.3$ & $+1.9$ & $+3.3$ & $+3.3$ & $+5.3$ & $+1.6$\\
    \bottomrule
  \end{tabular}
  \begin{tabular}{p{2cm} | p{0.52cm} p{0.52cm} p{0.52cm} p{0.52cm} p{0.52cm} p{0.52cm} p{0.52cm} p{0.52cm} p{0.52cm} p{0.52cm} | p{0.52cm}}
    \toprule
     & table & dog & horse & mbike & persn & plant & sheep & sofa & train & tv & mAP\\
    \midrule
    CNN-FV & $76.4$ & $82.0$ & $89.3$ & $86.1$ & $95.4$ & $65.0$ & $81.3$ & $70.8$ & $92.7$ & $76.7$ & $80.5$\\
    FisherNet & $\mathbf{78.1}$ & $\mathbf{88.0}$ & $\mathbf{92.2}$ & $\mathbf{88.3}$ & $\mathbf{96.9}$ & $\mathbf{67.2}$ & $\mathbf{81.7}$ & $\mathbf{74.1}$ & $\mathbf{94.7}$ & $\mathbf{78.0}$ & $\mathbf{83.4}$ \\
    \midrule
    Improvement & $+1.7$ & $+6.0$ & $+2.9$ & $+2.2$ & $+1.5$ & $+2.2$ & $+0.4$ & $+3.3$ & $+2.0$ & $+1.3$ & $+2.9$\\
    \bottomrule
  \end{tabular}
\end{table}

\begin{table}[t]
  \caption{The comparison of our FisherNet and traditional FV on PASCAL VOC 2012 using AlexNet (AP in $\%$). 
  Our learning strategy can boost the performance in all cases.}
  \label{table:improvement_2012}
  \centering
  \begin{tabular}{p{2cm} | p{0.52cm} p{0.52cm} p{0.52cm} p{0.52cm} p{0.52cm} p{0.52cm} p{0.52cm} p{0.52cm} p{0.52cm} p{0.52cm} | p{0.52cm}}
    \toprule
    & aero & bike & bird & boat & bottle & bus & car & cat & chair & cow &\\
    \midrule
    CNN-FV & $95.1$ & $81.1$ & $86.1$ & $83.5$ & $55.2$ & $86.8$ & $79.7$ & $91.0$ & $67.4$ & $74.3$\\
    FisherNet & $\mathbf{96.6}$ & $\mathbf{83.3}$ & $\mathbf{90.4}$ & $\mathbf{87.8}$ & $\mathbf{59.4}$ & $\mathbf{88.3}$ & $\mathbf{83.4}$ & $\mathbf{93.6}$ & $\mathbf{72.4}$ & $\mathbf{76.9}$ &\\
    \midrule
    Improvement & $+1.5$ & $+2.2$ & $+4.3$ & $+4.3$ & $+4.2$ & $+1.5$ & $+3.7$ & $+2.6$ & $+5.0$ & $+2.6$\\
    \bottomrule
  \end{tabular}
  \begin{tabular}{p{2cm} | p{0.52cm} p{0.52cm} p{0.52cm} p{0.52cm} p{0.52cm} p{0.52cm} p{0.52cm} p{0.52cm} p{0.52cm} p{0.52cm} | p{0.52cm}}
    \toprule
     & table & dog & horse & mbike & persn & plant & sheep & sofa & train & tv & mAP\\
    \midrule
    CNN-FV & $71.0$ & $88.2$ & $84.9$ & $88.4$ & $94.9$ & $58.5$ & $82.9$ & $63.0$ & $91.3$ & $79.3$ & $80.1$\\
    FisherNet & $\mathbf{75.3}$ & $\mathbf{91.2}$ & $\mathbf{89.5}$ & $\mathbf{89.1}$ & $\mathbf{96.8}$ & $\mathbf{61.1}$ & $\mathbf{83.3}$ & $\mathbf{66.7}$ & $\mathbf{93.1}$ & $\mathbf{80.9}$ & $\mathbf{83.0}$\\
    \midrule
    Improvement & $+4.3$ & $+3.0$ & $+4.6$ & $+0.7$ & $+1.9$ & $+2.6$ & $+0.4$ & $+3.7$ & $+1.8$ & $+1.6$ & $+2.9$\\
    \bottomrule
  \end{tabular}
\end{table}

\vspace{-0.37cm}
\section{Experiments}
\label{sec:exp}
\vspace{-0.37cm}

In this section, we will report our results for object classification, and do some discussions.

\vspace{-0.25cm}
\subsection{Experimental Setup}
\vspace{-0.25cm}
\label{sec:setup}

\paragraph{Datasets and Evaluation Metrics}

Two popular object classification datasets are chosen in our experiments, PASCAL VOC 2007 and 2012~\cite{Ref:Everingham2010}, 
which have $9962$ and $22531$ images respectively, with 20 different object classes. 
Both datasets have multiple labels for every image.
We train our model on the standard trainval set ($5011$ images for VOC 2007 and $11540$ for 2012) with only image level labels, and test on the test set. 
Average Percision (AP) and mean of AP (mAP) are used for evaluation.

\vspace{-0.25cm}
\paragraph{Implementation Details}
\label{sec:setup_detail}

As referred in Section~\ref{sec:pretrained_cnn}, our FisherNet is based on two CNN architectures AlexNet~\cite{Ref:Krizhevsky2012} and VGG16~\cite{Ref:Simonyan2015}, which are pre-trained on large scale dataset ImageNet~\cite{Ref:Deng2009}.

As the dimension of the second fc layer is $4096$, using it as patch features directly will result in high-dimensional FV. 
So we remove the last fc layer (mantain the first and second fc layers) and add two fc layers after the second fc layer: 
the first one is $256$-dimension for dimension reduction; 
the second one is for predicting image scores (its dimension is depended on the number of classes). 
Then we use the whole image to finetune this network on the target dataset. 
We train this network for $9000$ iterations with mini-batch size $32$. 
Learning rates of these two layers and other layers are set to $0.01$ and $0.001$ respectively, and divided by $10$ after every $3000$ iterations.
Results of this whole image finetuning procedure are shown as the CNN-finetune in Table~\ref{table:comparsion}.
After that, we use the SPP layer to replace the last max-pooling layer. Also, we replace the last fc layer of the finetuned model with our Fisher Layer and a fc layer for predicting scores.
We train the FisherNet $40000$ iterations, with mini-batch size $2$.
Learning rates of the Fisher Layer, the last fc layer, and other layers are set to $0.1$, $0.001$, and $0.0001$, respectively.
The number of GMM components is $32$, so the final dimension of image representation is $2\times32\times256=16384$.
The momentum and weight decay are set to $0.9$ and $0.0005$ respectively.

For Fisher Layer, we extract patch features from the $256$-dimension fc layer, then use GMM to get $\mathbf{\sigma}_{k}$ and $\mathbf{\mu}_{k}$ for initializing $\mathbf{w}_{k}$ and $\mathbf{b}_{k}$ in Section~\ref{sec:archi_fisher_layer}. 
Other new added layers are initialized using Gaussian distributions with mean $0$ and standard deviations $0.01$.

In all experiments, once our FisherNet is trained, we extract trained FV, and train a one-vs-all linear SVM with learning hyperparameter $C_{svm}=1$. 
As in \cite{Ref:Sanchez2013}, we use two normalization: power-normalization ($\mathbf{x} \gets sign(\mathbf{x}) \sqrt{\vert \mathbf{x} \vert}$) and L2-normalization ($\mathbf{x} \gets \mathbf{x} / \Vert \mathbf{x} \Vert_{2}$). 

\vspace{-0.25cm}
\paragraph{Other Setup}

To generate patches, we densely extract patches from seven scales $32 \times \{2, 3, ..., 8\}$, with step size $32$, which will produce $300$ to $800$ patches per-image. 
For data augmentation, we horizontally flip images for whole image finetuning. 
As our FisherNet can handle images with arbitrary sizes, we use five image scales $\{480, 576, 688, 864, 1200\}$ (resize its longest side to one of the five scales and maintain its aspect ratio) with horizontal flip for training FisherNet. 
Then we compute the mean of vector of these five scales (no flip) to train and test SVM. 

The GMM, SVM, and CNN are implemented by VLFeat~\cite{Ref:Vedaldi2010}, LIBLINEAR~\cite{Ref:Fan2008}, and Caffe~\cite{Ref:Jia2014}, respectively. 
All of our experiments are running on a PC with Inter(R) i7-4790K CPU (4.00GHZ), NVIDIA GTX TitanX GPU, and 32GB RAM.

\vspace{-0.25cm}
\subsection{Experimental Results}
\vspace{-0.25cm}

Experimental results on PASCAL VOC 2007 and 2012 are shown in Table~\ref{table:pascal_2007} and Table~\ref{table:pascal_2012}.\footnote{Results of our FisherNet on PASCAL VOC 2012 can be viewed at \url{http://host.robots.ox.ac.uk:8080/anonymous/DJ5JTS.html} and \url{http://host.robots.ox.ac.uk:8080/anonymous/AKKQXE.html}.} 
We can observe that our FisherNet achieves highly competitive results compared with other CNN based methods with single model. 
More importantly, our method outperforms other CNN-FV based methods~\cite{Ref:Liu2014,Ref:Cimpoi2015}. 
For example, Liu et al.~\cite{Ref:Liu2014} use outputs from the first fc layer as patch descriptors, and encode image using Sparse Coding based FV; 
Cimpoi et al.~\cite{Ref:Cimpoi2015} choose activations of the last conv layer as patch features, and extract patch features from ten different scales, then encode image using standard FV. 
These demonstrate the effectiveness of learning FV parameters and patch features. 

\vspace{-0.25cm}
\paragraph{Testing Time}

Our method is also very efficient. 
It takes only $0.3$s and $0.8$s per-image during testing, for AlexNet and VGG16 respectively, which is over $10\times$ faster than the previous state-of-the-art method HCP~\cite{Ref:Wei2015} ($3$s for AlexNet and $10$s for VGG16).

\vspace{-0.25cm}
\subsection{Discussion}
\label{sec:discu}
\vspace{-0.25cm}

Here we discuss benefits of our end-to-end training for object classification. 
Without loss generality, we only choose the AlexNet in this part.
Some results are shown in Table~\ref{table:comparsion}, where CNN-finetune means the whole image finetuning procedure as in Section~\ref{sec:setup_detail}; CNN-FV means extracting patch features using SPP, with the same patches as our FisherNet, then representing images by the standard FV; CNN-FL means setting learning rates before the Fisher Layer to $0$ and training our FisherNet, i.e., only learning FV parameters; FisherNet means our whole FisherNet training as in Section~\ref{sec:setup_detail}.\footnote{Some results on PASCAL VOC 2012 in Table~\ref{table:comparsion} can be viewed at \url{http://host.robots.ox.ac.uk:8080/anonymous/AN0JUF.html}, \url{http://host.robots.ox.ac.uk:8080/anonymous/38ZBIX.html}, and \url{http://host.robots.ox.ac.uk:8080/anonymous/RKWM6E.html}.}
We can observe that, the CNN-FV outperforms the CNN-finetune by a large margin, which demonstrates that BoVW based representation can achieve better performance compared with plain CNN.
The CNN-FL only achieves small improvements compared with the CNN-FV. Actually the CNN-FL is similar to \cite{Ref:Sydorov2014} which only learn FV parameters instead of learning parameters of FV and patch features jointly.
If we train all parameters jointly, the performance will be boosted a lot.
Also from Table~\ref{table:improvement} and Table~\ref{table:improvement_2012}, learning all parameters of patch features and FV jointly can boost the performance for all classes ($+2.9\%$ on PASCAL VOC 2007 and 2012 compared with the traditional FV). All these facts confirm the effectiveness of our learning strategy.

\vspace{-0.37cm}
\section{Conclusion}
\label{sec:conclu}
\vspace{-0.37cm}

In this paper, we propose a novel deep FisherNet framework, which makes all parameters of patch features and FV be learned in an end-to-end manner. 
Compared with traditional FV, experiments show substantial improvements by this learning strategy. 
We believe that using CNN based patch features with traditional BoVW based representation methods can achieve more satisfactory performance than plain CNN, and integrating these methods into a CNN framework can improve results further.
As FV is quite a effective image representation method and has achieved many state-of-the-art results on many computer vision applications like image classification and retrieval, in the future, we will explore how to use our new designed FisherNet for other applications.


\vspace{-0.37cm}

{
\small
\bibliographystyle{plain}
\bibliography{FisherNet}
}

\end{document}